# Explainable AI – the Latest Advancements and New Trends

Bowen Long, Enjie Liu S*Member,* Renxi Qiu, Yanqing Duan

*Abstract*— In recent years, Artificial Intelligence technology has excelled in various applications across all domains and fields. However, the various algorithms in neural networks make it difficult to understand the reasons behind decisions. For this reason, trustworthy AI techniques have started gaining popularity. The concept of trustworthiness is cross-disciplinary; it must meet societal standards and principles, and technology is used to fulfill these requirements. In this paper, we first surveyed developments from various countries and regions on the ethical elements that make AI algorithms trustworthy; and then focused our survey on the state of the art research into the interpretability of AI. We have conducted an intensive survey on technologies and techniques used in making AI explainable. Finally, we identified new trends in achieving explainable AI. In particular, we elaborate on the strong link between the explainability of AI and the meta-reasoning of autonomous systems. The concept of meta-reasoning is "reason the reasoning", which coincides with the intention and goal of explainable Al. The integration of the approaches could pave the way for future interpretable AI systems.

## I. INTRODUCTION

THE research around the theme of Trustworthy AI (TAI) is growing like never before, and the number of studies in the field of Trustworthy AI is growing by leaps and bounds. Since 2019, technology research around robustness, interpretability, and privacy protection has continued to grow [1]. Moreover, as the industry began implementing AI, the exploration and practice of making AI trustworthy continued to mature. The number of related AI technologies increased year by year after the integration of 'human-centric' computing elements.

The concept of Trustworthy AI began to gain attention around the mid-2010s, largely because growing concerns over the ethical, transparent, and safe deployment of AI systems. Standard organizations, such as IEEE introduced frameworks and standards, such as Ethics Certification Program for Autonomous and Intelligent Systems (ECPAIS) [2]. However, problems in the area of AI interpretability have slowed down due to the difficulty in cracking the black box attribute and the high cost of some of the methods for cracking the black box [3]. Lacking of interpretable AI techniques has become an important constraint on the fairness of the system [4]. Due to the lack of specific, well-developed scenario practice cases, and because related technology research is still in the early stages, there have been relatively few established patented in the areas of interpretability and fairness.

Current, industry proposes a shift from Trustworthy AI to more transparent AI. It not only requires technical support but also aims to create a systematic approach to generate more interpretable methods while maintaining high performance levels. The US Defense Advanced Research Projects Agency (DARPA) launched the Explainable AI (XAI) program [5] to create AI systems whose actions can be more easily understood by human users.

Although there has been a trend towards an explosion of research on this topic, not every area of it has received the same amount of attention. According to statistics from the China Academy of Communications Research [6], as of April 2022, patents in the field of privacy protection accounted for 63% of global AI technology patents, while AI interpretability and fairness patents accounted for only 10% and 6% respectively. With popularity of AI and increasing applications of AI-enabled applications, trustworthiness and transparencies of AI becomes an unavoidable issue in AI development, and hence we witness the increasing development in the area.

In this paper, we first surveyed developments from various countries and regions on the ethical elements that make AI algorithms trustworthy; we adopted the definition from EU AI High-level Expert Group (HELG) [7], which summarized four basic ethical principles; AI HELG also identified seven specific needs to meet the ethical principles. We focus on one of the main requirement of the ethical AI, which is the transparency. To be more specific, we focus on reviewing of the latest research on improving explainability of AI. Finally, we also present our thoughts on the future development of technologies in AI explainability, and possible techniques to realize it.

- Bowen Long, Enjie Liu, Renxi Qiu, are from School of Computer Science and Technology, University of Bedfordshire, Park Square, LU1 3JU, Luton, UK
- Yanqing Duan is from Business School, University of Bedfordshire, Park Square, LU1 3JU, Luton UK.
- E-mail: (bowen.long, enjie.liu, renxi.qiu, yanqing.duan)@ beds.ac.uk
- Enjie Liu (Enjie.liu@beds.ac.uk) is the corresponding author

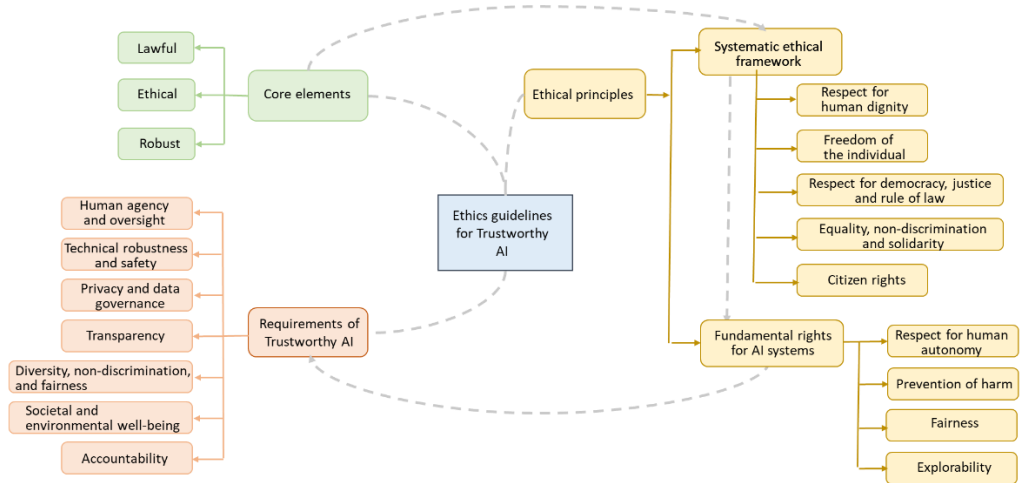

Fig. 1 Core elements with the ethics guidelines for trustworthy AI

## II. ETHICAL CONCERNS OF TRUSTWORTHINESS IN AI

European Commission proposed civil rules on robotics in [8], through two public statements in April and December 2018 [9], it clearly articulated three main visions for the field of Artificial Intelligence:

- to increase investment from all sectors of society in the development of AI;
- to prepare for new economic trends brought about by AI;
- to ensure that appropriate ethical and legal frameworks are in place so that European values in AI remain centered around the human being.

To achieve these common goals, the EU convened 52 authoritative experts or scholars, as well as authoritative practitioners to form the AI HLEG, which brought them together to draft and revise the Code of Ethics for Artificial Intelligence [13]. In this official document, the concept of trustworthy AI was formalized.

With due foresight and a sense of responsibility, the AI HLEG may have anticipated that AI governance would require more than merely outlining a vague concept of Trustworthy AI in broad terms. They have defined Trustworthy AI, through: (1) the core elements it should have; (2) the ethical principles it should protect; and (3) the specific requirements it should fulfil. These three aspects have been explored in depth to find ways to prevent AI from causing a crisis of trust [6], as shown in Fig. 1. Table 1 summarizes the investigation and progress of ethical concerns from different countries and regions.

| Where | When | Documents/projects | Elements |
|---|---|---|---|
| The United States | 10/2019 | Principles of Artificial Intelligence: Ethical Recommendations for the Application of Artificial Intelligence in the Ministry of Defence [10] | 1) accountability; 2) fairness; 3) traceability; 4) reliability; and 5) control |
| | 02/2020 | "Ethical Standards for AI Guidance." [11] | Conduct AI system development, testing and review to the highest standards of fairness, accountability and transparency |
| | 07/2021 | "Trustworthy and Responsible" Artificial Intelligence Project [12] | 1) Accuracy; 2) Interpretability; 3) Privacy; 4) Reliability; 5) Robustness; 6) Safety; 7) Reducing the harm of bias |
| The European Union | 01/2017 | European Commission Recommendation on Civil Law Rules on Robotics [9] | 1) freedom; 2) privacy; 3) positive value and dignity; 4) self-determination and non-discrimination; 5) personal data protection |
| | 04/2019 | Ethics guidelines for trustworthy AI [13] | 1) human oversight; 2) robustness and safety of technology; 3) privacy and data management; 4) transparency; 5) diversity, non-discrimination and fairness; 6) social and environmental well-being; 7) accountability |
| Japan | 03/2017 | Ethical Guidelines of the Society for Artificial Intelligence [14] | 1) Contribution to humanity; 2) Compliance with legal rules; 3) Respect for the privacy of others; 4) Impartiality; 5) Safety; 6) Honest behaviour; 7) responsibility to society |
| | 07/2017 | AI Development Programme | Developers should endeavour to exclude, to the extent possible, all measures of bias and other undue discrimination, including learning data from AI systems, in light of the nature of the technology; developers should comply with international human rights law, international humanitarian law, and be on the lookout for behaviour by AI systems that maybe incompatible with human values. |

| | 03/2019 | Principles for a Human-Centred AI Society | 1) Human-centred; 2) Educational applications; 3) Privacy Protection; 4) Safety and Security; 5) Fair Play; 6) Equity; 7) Accountability and Transparency; 8) Innovation |
|---|---|---|---|
| Canada | 12/2018 | Montreal Declaration on Reliable Draft Artificial Intelligence | 1) well-being; 2) autonomy; 3) justice; 4) privacy; 5) knowledge; 6) democracy; 7) responsibility |
| Australia | 11/2019 | Australian Ethical Framework for Artificial Intelligence | 1) human, social and environmental well-being; 2) human-centred values; 3) fairness; 4) privacy protection and security; 5) reliability and safety; 6) transparency and interpretability; 7) contestability; 8) accountability |
| New Zealand | 03/2020 | Guiding Principles for Trustworthy AI in New Zealand | 1) Fairness and Justice; 2) Reliability, Security and Privacy; 3) Transparency; 4) Human Oversight and Responsibility; 5) Welfare |
| Korean | 12/2020 | National Ethical Standards for Artificial Intelligence | 1) Human Rights Protection; 2) Privacy Protection; 3) Respect for Diversity 4) Prohibition of Infringement; 5) Social Advocacy; 6) Cooperation of Subjects; 7) Data Management; 8) Clarification of Responsibilities; 9) Ensuring Safety; 10) Transparency |
| China | 11/2017 | S36 Xiangshan conference | Trustworthy AI consists of three main elements: human, information, and physical. |
| | 08/2019 | Artificial Intelligence Industry Self-Regulation Convention | 1) Safe and Controllable; 2) Transparent and Releasable; 3); Protecting Privacy; 4) Clear Responsibility; 5) Diverse and Inclusive |
| | 09/2019 | Code of Ethics for New Generation Artificial Intelligence | 1) Enhance human well-being; 2) Promote fairness and justice; 3) Protect privacy and security; 4) Ensure controllability and credibility; 5) Strengthen responsibility; and 6) Enhance ethical literacy. |
| | 07/2021 | White Paper on Trusted Artificial Intelligence. | Trustworthiness consists of five trustworthy elements: 1) reliability and control, 2) transparency and release, 3) data protection, 4) clear responsibility, and 5) diversity and inclusiveness. Trustworthy AI is the implementation of ethical governance requirements from the perspective of technology and engineering practice to achieve an effective balance between innovation development and risk governance. |

Table 1. Summary of ethical concerns from different countries (Translated from [6])

Recently, after surveying the principles of the European Group on Ethics in Science and New Technologies (EGE) as well as 36 other ethical principles proposed so far, the AI4People task force summarized them into nine main principles [15]. Subsequently, the AI HELG, in response to these, refined them and selected four as the basic ethical principles that trustworthy AI should always adhere:

- Respect for Human Autonomy
- Prevention of Harm
- Principle of Fairness
- Principle of Explicability

Abstract ethical principles need to be translated into tangible operational requirements to establish a solid foundation. AI systems cannot be realized without the involvement of relevant stakeholders: a.) developers who implement and apply them to the development process; b.) developers who ensure that the system and products and services comply with the requirements; and c.) users who are willing to use them and give feedback. In view of this, AI HELG has summarized seven specific needs to meet the ethical principles in place by reflecting on the stakeholders' positions, shown in Fig. 2.

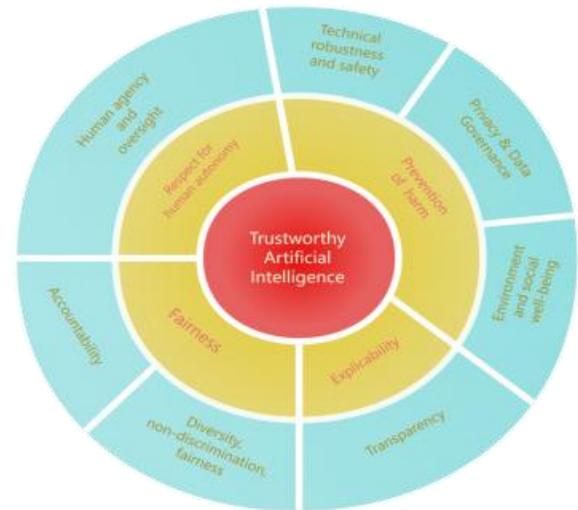

Fig. 2 Relationship between the seven requirements and the corresponding basic principles

III. TRANSPARENCY AI, INTERPRETABLE AI, AND EXPLAINABLE AI

The Trustworthy AI is an umbrella term for concepts that make AI socially valuable when utilized and deployed. Transparent AI is one of seven requirements that trustworthiness should be met. Research on transparent focuses on the openness of structure and processes in AI models. Research on Interpretable AI focuses on understanding the outputs of AI systems. Explicability is critical in building and maintaining user trust in the system. When AI systems can be explained, it is transparent, and therefore, the model is trusted. However, a system can be transparent but not interpretable.

### A. Transparency of AI

Transparency includes traceability, interpretability and communication.

#### a) Traceability

It means that the datasets and processes the system relies on to make decisions—including data collection methods, reasons for data labeling, and the algorithms used—should be documented as thoroughly as possible. To achieve traceability and increase transparency, this also applies to decisions made by AI systems. Traceability facilitates review and interpretation.

#### b) Interpretability

It involves the technical process of explaining the AI system and it requires that the decisions taken by the AI system can be understood and tracked by human. In addition, there may be trade-offs between increasing the interpretability of a system or increasing its accuracy [16]. Whenever an AI system has a significant impact on people's lives, individuals should be able to demand a proper explanation of the decision-making process. In addition, the extent to which the AI system affects the organization's decision-making process, along with the rationale for its design and deployment, should be explained.

#### c) Communication

Humans have the right to be informed that they are interacting with an AI system. This requires the AI system be recognizable. In addition, the basic right to decide not to interact with the AI system and to switch to interacting with humans should be provided if necessary. In addition to this, the performance and limitations of the AI system should be communicated to the AI practitioner or end user in a suitable manner.

### B. Interpretable AI and explainable AI

*Interpretable AI* and *Explainable AI* are sometimes interchangeably used, but there are distinct meanings: *Interpretable AI* is often used in designing models that are inherently understandable, it makes a model understand from design stage; *Explainable AI*, on the other hand, is often referred to the techniques and methods to make the results understandable.

*Interpretable AI* is one of the three attributes of Transparent AI, which is intended to indicate what a particular area of AI is trying to achieve. Interpretability of AI started as early as the mid-1970s [17].

*Explainable AI* was first introduced in 2004 [18], it involves different perspectives such as social science, cognitive science, psychology, and human-computer interaction, and aims to make the results of AI systems more understandable to humans. In [19], the authors summarized interpretability approaches used in machine Learning.

In some studies, *Interpretable AI* is categorized as a subfield of *Explainable AI*, which is used to express the technical behavior in the process of making AI systems explainable.

In this paper, we do not intend to distinguish the difference between the interpretable AI and explainable AI in our survey. We aim to survey the state of the art in explaining why a model produces a particular output or decision. With the so-called black box, multiple interpretable measures may be required to collaborate, and the degree of interpretability needed depends very much on the circumstances. In this paper, we categorize the interpretive approach as range-based, where interpretation is achieved globally or locally, and sequence-based, where interpretation is achieved at different stages.

## IV. RANGE-BASED INTERPRETIVE APPROACH

There are two different forms of interpretability, (1) understanding the behavior of the whole model - global interpretability: and (2) understanding individual regional predictions - local interpretability.

### A. Global Interpretability

Global interpretability helps users understand the overall logic of the model. To interpret black-box machine learning models globally, the authors in [20] used a compact binary tree and an interpretation tree to explicitly represent the most important decision rules implicitly contained within the black-box models. In such cases, global explanations are more helpful than explanations made for a particular segment. Authors in [21] proposed an a priori learning technique in the form of a deep generator network based on method of activation maximization [22]. However, global model interpretability is difficult to achieve in practice for models with a small number of parameters.

### B. Local Explainability

Local explanations are the most common method for generating Deep Neural Network (DNN) explanations. It explains the reasons for a particular decision or a single prediction implies that the interpretability occurs locally. The mining of interpretability is used to generate individual explanations and, in general, to justify why a model makes a given decision in a particular situation. Authors in [23] proposed the Locally Interpretable Model agnostic Explanatory (LIME) model. This model can locally approximate a black-box model in the vicinity of any prediction of interest. Authors in [24] proposed a new technique called Shapely Explanations based on the research of [25]. In addition, limited by the fact that global explanations are not well explained, researchers have tried to link local explanations with holistic explanations [26-28].

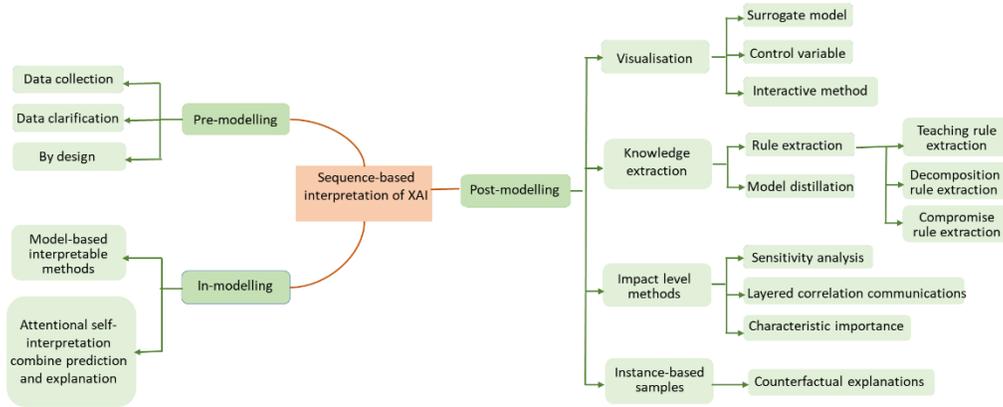

Fig. 3. Summary of the sequence-based interpretative approaches

## V. SEQUENCE-BASED INTERPRETIVE APPROACH

Through a review of the past literature, study found that although many literatures have proposed a classification of interpretive methods [29], there is always a lack of uniformity. For the interpretive methods in Trustworthy AI, this study adopts the following two scales: (1) based on the scope, as explained in the previous section; (2) based on the sequence, which will be explained in this section. Fig. 3 summarizes the sequence-based interpretive methods in explainable AI.

The interpretability of AI can occur at three stages: pre-modelling, in modelling, and post-modelling.

### A. Pre-modelling

Pre-modelling interpretability means that it occurs before the model development process, and this stage mainly involves data collection, data classification, and model a priori design of the model.

*(1)   Data Collection* Many researchers use data augmentation for data complexity addition. In a series of works [30–32] and others, a new sample dataset was generated for training by applying a series of rotations, flips, and noise additions to the images using self-supervised learning.

*(2)   Data classification* Authors in [33] proposed the prototype network to measure data similarity and was widely adopted. Authors in [34] tackled the few-shot learning problem from a multimodal (text and image) perspective. The paper proposed to leverage a nearest neighbor classifier in a powerful representation space. Classification is performed by finding the nearest multimodal class prototype to an unseen test sample. However, the uniqueness of the prototype points within each data class leads to test sample classification results being determined by their similarity to a prototype point. It may result in over-generalization or overfitting leading to serious data bias. In [35], authors extended the approach, by proposing a discriminative sample generation along with a self-paced strategy for sample selection a subset of "easy" samples can be automatically selected in each iteration. Training is performed using only subset, which is progressively increased in the subsequent iterations when the model becomes more mature. Authors in [36] attempted to use semantically-assisted visual classification, but the area of visual-to-semantic is yet to be demonstrated.

*(3)   By design* In general, the more complex a model is, the more difficult it is to explain it. Therefore, the best way to obtain an interpretable AI model is to find an algorithm that is as interpretable as possible at the model setup expense.

Authors in [37] proposed a model called Bayesian Rule Lists (BRL) that based on BRL of Decision Trees in the hope of gaining the trust of relevant practitioners through its simplicity and convincing ability. Authors in [38] introduced a generalized summation based model to seek a solution to the problem of pneumonia. Authors in [39] introduced an attention based model that automatically learns to describe the content of an image. Authors in [40] proposed a linear model to create a data-driven scoring system. Authors in [41] has suggested that "accuracy usually requires more sophisticated prediction methods and simple interpretable functions do not make the most accurate predictors"

As explained in section IV, explainability can be categorized into two main paradigms: global explanation and local explanation. Table 2 summarises method used in pre-modelling for achieving interpretable AI under these two categories.

### B. In modelling

Interpretability in modelling refers to the development of measures that inherently explain the model. At this stage, researchers utilized model-specific explanatory methods, prediction and explanation of attentional self - interpretation for joint modelling.

*(1)   Model-specific interpretable methods* This approach is model-specific and model-independent, which has led to a surge of interest. Authors in [13] and others have investigated gradients in convolution and inverse convolution to explore relationships in convolutional neural networks, and authors in [42] have simplified the structure of convolutional neural networks to a fully convolutional structure for the task.

| Range-based | Method | Conditions of application | Technology/Algorithm/Model | Application Scenarios |
|---|---|---|---|---|
| global | Interpretable Models | Transparent parameters | Explainable Decision Trees [43] | Self-driving |
| | | | Explainable Linear Regression [44] | Crime forecasting |
| | | | Explaining Logistic regression [45] | Human resources management |
| | | | Explaining Bayes' rule based algorithms [456 | Spam filter |
| | | | Explainable Support Vector Machines (SVM) [47] | Manufacturing quality control |
| | Statisticians | Prior knowledge and experience | Causal Fuzzy Cognitive Maps [48] | Urban planning and governance |
| | | Sensor-based | Spike Neural Networks (SNN) [49] | Robot Sensing and Control |
| Local | Interpretable Models | Transparent parameters | Gradient Boosting Decision Trees(GBDT) [50] | E-commerce search engine sorting |
| | Statisticians | Probabilistic and causal | Structural Causal Models [51] | Environmental science research |
| | | | Bayesian Causality Networks [52] | Production process optimisation |
| | Counterfactual Explanations | Data-based | Generative Adversarial Networks (GAN) [53] | Data enhancement and sample generation |
| | | Scenario-based | Diverse Counterfactual Explanations (DiCE) [54] | Personalised Recommender System |
| | Agnostic | Data-based | Explainable Principal Component Analysis (PCA) [55] | Financial analytics and customer behaviour analysis |
| | | | t-Distributed Stochastic Neighbour Embedding (t-SNE) [56] | Analysing similarities in graphic-textual data |
| | | | Uniform Manifold Approximation and Projection (UMAP) [57] | Analysing biogenetic data |

Table 2 Technologies used in prior modelling of explainable AI

*(2)    Attentional self-interpretation* It combines prediction with interpretation. By assigning weights and importance in the data, attention-based approaches are considered more promising. The authors in [58] explored the Visual Question Answering (VQA) task of answering questions about images in free-form natural language, where the VQA model incorporates textual information during the question-and-answer process. The authors in [59] proposed fault and knowledge prediction metrics and found that the HIL (Human-In-Loop) approach enables human understanding. Authors in [60] extended the approach of [61] by interpreting the results of a VQA task by using textual interpretations and the visual regions corresponding to them. This process passes an attentional mask between modules and explores inter-modal complementarity to demonstrate the value and advantages of generating multimodal interpretations. A supervised attentional model was used in the study [62] in order to train the generation of interpretations that resemble human reasoning.

Table 3 summarizes the latest development of in-modelling approaches that attempted for achieving expansible AI.

*C. Post-modelling*

Post-modelling approaches, also often referred to as model-independent or post-hoc interpretation approaches [66-67] are currently a frequently cited classification approach in the field of XAI. In this study, we draw on the model-independent categorization of XAI in [68] and others, and set the classification catalogue as: visualization, knowledge extraction, influence methods, and factual descriptions.

*(1)    Visualization*

Visualization techniques are mainly applied to supervised learning models and are one of many effective methods for exploring artificial intelligence systems. The elements introduced in this study are categorized as follows:
- *Surrogate models*
- *Control variable graphs*
- *Interactive methods*

a. Surrogate model

It is trained from the predictions of the original black box model. It can be used to explain complex models. The LIME [67] method is a prescribed way to build a local surrogate model around a single observation. The authors in [69] used a surrogate model approach to extract a decision tree representing the behavior of the model. Another study by [70] proposed a method of using surrogate models for constructing a tree view visualization method.

Table 3 Technologies used in in-modelling of explainable AI

| Range-based | Conditions of application | Technology/Algorithm/Model | Application Scenarios |
|---|---|---|---|
| global | Simplification-based | Architecture modification techniques [63] | Signal enhancement for speech recognition |
| global& local | Parameters-based | Backpropagation-based methods [64] | Significant characterisation of imaging tasks |
| local | Weighting-based | Attention mechanisms [65] | Natural Language Processing (NLP) |

b. Control variables

This method refers to the establishment of a link between outcome and cause by showing the average partial relationship between one or more input variables and the predictions of the black box model. Partial Dependence Plot (PDP) and Individual Conditional Expectation (ICE) are used in express the relationship between the outcomes and the causes. Partial Dependence Plot is a method that uses a graphical representation of the relationship between an objective function and a feature, which helps to visualize the overall effect of variable changes on the results. The study of Partial Dependence Plot [71], showed the marginal effects of the final predictor features. The authors in [72] demonstrated the advantages of using Random Forests and correlation PDPs to accurately model the predictor-response relationships at asymmetric categorical costs in a criminal justice setting. Individual Conditional Expectation reveals interactions and individual differences based on PDP graphs. Authors in [73] presented a preliminary theory of ICE and elucidated the advantages over PDP. The authors in [74] and proposed representation based on local feature importance.

c. Interactive methods

Interactive methods were initially designed to exploit the transparency of machine learning to analyze the accuracy of multimodal systems [75], but have since been commonly used for interpretability studies. Authors in [76] applied a virtual agent to improving user trust, finding that multimodal explanations of the combination of vision and speech were more convincing. The authors in [77] proposed an interactive algorithm that was used by XAI go to explain the internal state of the system by answering questions now. The authors in [78] introduced an interactive model with an "active attention" mechanism. By combining model interpretation and annotation, model interpretation is evaluated based on metrics such as user trust, mental models, and usability. This approach adjusts model attention and provides user feedback to correct inaccurate predictions, ultimately improving accuracy. Interpretation-based dialogue systems provide users with better explanations than traditional systems by using an interactive approach that integrates multiple senses [79].

*(2)* Knowledge Extraction

The task of extracting knowledge from the network is to extract the knowledge gained by the neural network during training in an understandable form and encode it into an internal representation. There are two main methods designed to extract knowledge from artificial neural networks in this study: rule extraction and model refinement.

a. Rule Extraction

To gain insights into highly complex models, many researchers have begun experimenting with rule extraction [80]. A symbolic and comprehensible description of the knowledge learned by the network during the training process is provided by extracting rules, which approximate the decision-making process of the AI network by utilizing its inputs and outputs. In a study [81], rule extraction strategies are classified and three rule extraction models are proposed: pedagogical rule extraction, decomposition rule extraction and compromise rule extraction.

b. Model distillation

Distillation is a type of model compression that transfers knowledge learned by the 'teacher' network (a deep network) to the 'student' network (a shallow network) [82, 83]. Model compression was initially used to reduce the computational cost of running a model, but has since been applied to interpretability as well. The authors in [84] investigated how model distillation can be used to distill complex models into forms that are as transparent as possible. The authors in [85] introduced a new method of knowledge distillation, known as the Interpretable Imitation Learning (IIL) method. Since then, research in [86-87] have expanded the approach.

*(3)* Impact level methods

This type of approach estimates the importance as well as the relevance of features or modules by changing inputs or internal components and recording the trend in the degree of impact of these changes on the model performance. Relevant methods are: sensitivity analysis, hierarchical correlation propagation, and feature importance metrics.

a. Sensitivity analysis

It is used to verify that model behavior and outputs remain stable when the data is deliberately perturbed or when other changes occur in the simulated data [88]. Sensitivity analysis has been increasingly used to explain image classification in DNNs [89-90].

| Range-based | Method | Conditions of application | Technology/Algorithm/Model | Application Scenarios |
|---|---|---|---|---|
| Post-hoc (local) | Agnostic | Data-based | Local Interpretable Model-Agnostic Explanations (LIME) [98] | Data science experiments |
| | | | Perturbation-based methods [99] | Assessment of insurance claims |
| | | Feature-based | SHapley Additive exPlanations (SHAP) [100] | Climate change projections |
| | | | Feature Importance Scores [101] | Software for evaluating AI models |
| | | | Attribution-based methods [102] | Evaluation of corporate business indicators |
| | | | Local-perturbation techniques [103] | Comparison and classification of computer vision. |
| | | | Sensitivity/Stability-based methods [104] | Sentiment Analysis in NLP |
| | | Graph Example-based | Graph Neural network explainer (GNN Explainer) [105] | Knowledge Reasoning |
| | | | Gradient-Based Techniques: Class Activation Mapping (CAM) [106] | Anomaly Detection in Security Monitoring |
| Post-hoc (global) | Specific | Feature-based | Feature contribution techniques [107] | Trend analysis of house prices |
| | | Simplification-based | Activation mapping [108] | Visualisation of target detection |
| | | Surrogate | Model distillation [109] | Cloud Computing Tasks |
| | Statisticians | Probabilistic and causal | Explainable Naive hierarchical Bayes models [110] | Recommender system |
| | | | Bayesian Belief Networks [111] | Troubleshooting of industrial equipment |
| | | Feature-based | Partial Dependence Plot (PDP) [112] | Financial risk management |
| | | | Regression Concept Vectors (RCV) [113] | Medical Diagnostics |

Table 4 Technologies used in post modelling of explainable AI

b. Layered Correlation Propagation

Layered Correlation Propagation Algorithm (LRP) was first proposed in [91]. The algorithm performs a backward redistribution of the prediction function, starting at the output layer of the network and then back-propagating to the input layer. The key property of this redistribution process is known as correlation conservation.

c. Characteristic importance

Characteristic importance quantifies the contribution of each input variable to the prediction of a complex ML model. The authors in [92] used a quantitative ranking of feature importance when proposing a decision boundary based model to explain the data classification, calculating the increase in model prediction error can be used to measure the importance of features. Similarly, [93] proposed a model-independent version of feature importance called Model Class Dependency (MCR). The authors in [94] proposed a localized version of feature importance, permutation-based Shapley's Feature Importance (SFIMP) to fairly distributes the model performance among features and compares feature importance across different models.

*(4)* Instance-based samples

Most example-based explanations are not directly related to the model itself; sample-based explanation methods explain the model by selecting representative samples from the dataset. A common approach is *counterfactual interpretation*, which reasons by assuming the opposite of the predicted outcome. Contrastive, human-based reflection on failed events can prompt people to think about why a machine arrives at a particular result rather than other possible outcomes. For example, suppose a loan application is rejected [95]. In this case, there are questions about which steps and minimal changes could have been made to alter the undesirable outcome, and in [96], authors proposed the concept of an unconditional counterfactual explanation as a new approach to automated decision making. Multi-modal explanations based on counterfactuals can provide actionable insights and recommendations [97].

Table 4 summarizes the latest development of post-modelling approaches that attempted for achieving expansible AI.

VI. NEW TRENDS IN EXPLAINABILITY OF AI, OPPORTUNITIES AND CHALLENGES

*A. Explainability AI and meta-reasoning*

A key challenge with the sequence-based and range-based methods surveyed in this paper is that explainability is obscured by the complex interactions between the learning and reasoning, both are essential requirements of AI systems, making it extremely difficult to extract. We advocate that the explainability of AI systems can be approached by projecting the problem into the reward space for a reward-driven explainability. The approach focuses on explaining the potential impact of the AI systems using logical reasoning techniques solely at the reward space, which significantly reduced complexity and improves observability.

The concept of meta-reasoning is "reason the reasoning" [136], which coincides with the intention and goal of explainable AI. The integration of the approaches could pave the way for future interpretable AI systems.

Meta-reasoning (as shown in Fig. 4) aims to simplify the symbolic grounding processes by observing patterns of reward for explanation rather than sequence-based or range-based causal relationships. Traditionally, logical reasoning in AI has been used for object driven knowledge representation and reasoning but it is often inadequate at ground level required by data driven models. Meta-reasoning consists of the meta-level control of computational activities and the introspective monitoring of reasoning [114]. The concept has been successfully applied in Neuroscience under the umbrella of Meta-analysis for the synthesis of quantitative data from multiple independent models [115].

Compared to meta-learning, which focuses on the efficient use of data at the ground level, meta-reasoning emphasizes the efficient use of computational models and resources at the object level, which is crucial for the trustworthiness of AI systems. We advocate that the achieving explainability of AI systems by projecting the problem into the reward space for a reward-driven explainability is a new research trend. This simplification avoids the data-driven chaotic systems that are difficult to define with the reductionist view expected by trustworthy AI [116, 117].

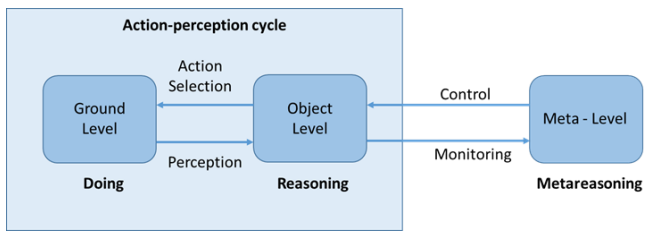

Fig. 4 Duality in reasoning and acting

The reasoning and explainability of AI share a common goal at the meta-level: at the object level, to explain the observation of the ground level (action-perception). The goal of meta-level control is to improve the quality of its decision-making by determine what and how much reasoning to perform, rather than focusing solely on actions. The goal of the explainability is to verify of the model consistent with its design.

In [118], authors identified variants of meta-reasoning. Suppose there is a fixed set of possible plans of action (decision) that the agent can choose from, the agent has some extant expected for each plan. The agent can take some deliberation actions on each plan; depending on the rewards of the deliberation action, the expected utility of that plan changes. The goal here is to find a deliberation strategy that maximizes the expected utility/reward of the agent. Another variant is that the agent knows that there are multiple states in clarification/prediction, to obtain nonzero utility, the agent needs to determine (by deliberation) the state with certainty.

*B. Trustworthy autonomous systems and domain randomization*

The main challenges in achieving explainability involve establishing a mathematical framework for monitoring and control at the meta-level. Bayesian based networks allow us to learn a probability distribution over possible behaviors of neural networks, bridging the information between the ground levels and object levels. To do so, techniques such as *trustworthy autonomous systems* and *domain randomization* are integrated to create more explainable AI systems at the reward space. Trustworthy autonomous systems and domain randomization are not inherently "reward-based" by definition, but they can involve reward-based mechanism.

Trust is a psychological state that can be established through interactions with corresponding agents. Trust in AI can be built by interacting with autonomous systems. Given that the AI in trustworthy autonomous systems [119] is directly exposed to the reward space, its explainability can be observed and verified by explaining the behavior of the corresponding autonomous systems according to their design. Meanwhile, domain generalization enables AI to make decisions with less reliance on a distributed training area. It aims to learn a generalizable model from multiple known source domains and then directly generalize to an unseen target domain without additional adaptation. This approach has been widely studied for creating more robust neural networks by transferring source images to different styles in spatial space to learn domain-agnostic features. This significantly reduces the impact of network errors, allowing explainability to focus solely on the real rewards generated by ground-level information.

Domain randomization has been used in robotic training [120-124] and image clarification [125-127] the required real-world data is typically prohibitively expensive to acquire, or fragile to external disturbance. Unfortunately, such polices are often not transferable to the real world due to a mismatch between the simulation and reality, called 'reality gap'. Domain randomization methods address this problem by randomizing the physics simulator (source domain) during training according to a distribution over domain parameters in order to obtain more robust policies capable of overcoming the reality gap. It manipulates the training data according to rewards e.g. learning in some specific simulations or deliberately adding random noise in the training data could improve the robustness of the neural network.

In [128], the authors proposed style randomization, a feature-level augmentation strategy, can increase the networks' generalization capability simply by diversifying the source domains. In [129], authors proposed a Variance Reduced Domain Randomization (VRDR) approach for policy gradient methods, to strike a tradeoff between the variance reduction and computational complexity for the practical implementation. In [130], the authors proposed a single domain generalization for remote sensing image segmentation via domain randomization of texture and style information. In [131], the authors proposed an unsupervised cross-domain object detection method based on multi-domain randomization. Domain randomization parameter callback module is devised to retain the key detection information of the object, thereby improving the model's stability. A source domain consistency loss is incorporated to enhance the convergence speed of the model and amplify the semantic information embedded within the features.

*C. Large Language Models in explainable AI*

Recent advancements in Large Language Models (LLMs) have significantly enhanced the explainability of AI through automated and structured reasoning capabilities. The exceptional abilities of LLMs in natural language understanding and generation are vital for orchestrating meta-reasoning, which helps identify the most appropriate explanations for AI systems based on observations. This rapidly growing field has seen substantial progress over the past year and is poised to play a crucial role in various approaches to achieving explainable AI.

The authors in [132], [133], and [134] use Chain-of-Thoughts and Tree-of-Thoughts to convert LLMs into reasoners, thereby explaining the potential of AI from a problem-solving perspective. While these methods enhance explainability in targeted scenarios, they fall short of achieving consistent performance across different tasks [135]. This limitation can be mitigated by meta-reasoning, which ensures the monitoring and control of the reasoning paradigm at the meta level [116]. Therefore, a key advantage of integrating meta-reasoning with LLMs is the ability to adjust strategies based on the context and specific task requirements. This is essential for addressing the diversity and complexity of real-world problems and has been applied to Meta-Reasoning Prompting (MRP), which endows LLMs with adaptive explaining capabilities [135]. the authors in [137] proposed a novel approach, which to integrating Shapley Additive exPlanations (SHAP) values with LLMs to enhance transparency and trust in intrusion detection. In [138], the authors combine Markov logic networks (MLNs) with external knowledge extracted using LLMs, aiming for improved both interpretability and accuracy. In [139], domain-experts were integrated in the prompt engineering, as man-in-the-loop, authors designed the explanation as a textual template, which is filled and completed by the LLM. [140] compared global rule-based models (LLM and DT) with well-established black box models. Rule-based models has simpler shapes than the ones resulting from more complex black-box models like Support Vector Machine (SVM) or Neural Networks.

VI. CONCLUSION

In this paper, first surveyed the state-of-the-art standards and developments from various countries and regions on the ethical elements that make AI algorithms trustworthy. We then surveyed the techniques used in achieving explainable AI. Lastly, we identify the new trends in addressing explainability, and also listed potential solutions.